\documentclass[conference]{IEEEtran}
\IEEEoverridecommandlockouts
\usepackage{cite}
\usepackage{amsmath,amssymb,amsfonts}
\usepackage{algorithmic}
\usepackage{graphicx}
\usepackage{textcomp}
\usepackage{xcolor}
\usepackage{balance}
\def\BibTeX{{\rm B\kern-.05em{\sc i\kern-.025em b}\kern-.08em
    T\kern-.1667em\lower.7ex\hbox{E}\kern-.125emX}}

\begin{document}

\makeatletter
\def\input@path{{./files/}}

\graphicspath{{./figures/}}

\title{MoniLog: An Automated Log-Based Anomaly Detection System for Cloud Computing Infrastructures}

\author{\IEEEauthorblockN{Arthur Vervaet (2023-10-31)}
\IEEEauthorblockA{\textit{Supervised by Professor Raja Chiky \& Mar Callau-Zori} \\
\textit{ISEP - Institut Supérieur d’Électronique de Paris / 3DS OUTSCALE}\\
Paris, France \\
arthur.vervaet@outscale.com}
}

\maketitle

\begin{abstract}
  Within today's large-scale systems, one anomaly can impact millions of users.
  Detecting such events in real-time is essential to maintain the quality of services.
  It allows the monitoring team to prevent or diminish the impact of a failure.
  Logs are a core part of software development and maintenance, by recording detailed information at runtime.
  Such log data are universally available in nearly all computer systems. 
  They enable developers as well as system maintainers to monitor and dissect anomalous events.
  For Cloud computing companies and large online platforms in general, growth is linked to the scaling potential.
  Automatizing the anomaly detection process is a promising way to ensure the scalability of monitoring capacities regarding the increasing volume of logs generated by modern systems.
  In this paper, we will introduce MoniLog, a distributed approach to detect real-time anomalies within large-scale environments.
  It aims to detect sequential and quantitative anomalies within a multi-source log stream. 
  MoniLog is designed to structure a log stream and perform the monitoring of anomalous sequences.
  Its output classifier learns from the administrator's actions to label and evaluate the criticality level of anomalies.
\end{abstract}


\section{Introduction}

Cloud computing platforms give access to on-demand IT resources.
This outsourcing makes Cloud providers responsible for the high availability and quality of their client services.
Within such large-scale online systems, one incident can impact millions of users\cite{liang2020robust, liu2019fluxrank, wang2020practical}.
Anomaly detection is a critical step towards building secure and trustworthy platforms.
Timely and accurate detection of anomalous sequences enables operational teams to mitigate their losses\cite{zhang2018prefix}.

Recording runtime information is a common practice within software systems \cite{li2019generic}. %barik2016bones
The produced logs describe a vast range of events as well as variations in the system states.
Because of its simplicity and effectiveness, logging has been commonly adopted in practice\cite{zhu2015learning}.
Logs are therefore one of the most valuable data sources for anomaly detection \cite{satpathi2019learning, khatuya2018adele, zhang2019robust}.

Maintainers and developers use logs to understand the status of a system, identify the origin of breakdowns
\cite{zhu2019tools,fu2013contextual, kobayashi2017mining}, dissect performance issues \cite{jha2020live,lu2020cloudraid},
and prevent security attacks\cite{moh2016detecting}.

Large-scale service and its underlying systems development generally involves hundreds of human actors divided into different teams and sometimes different structures.
Developers/operators usually have incomplete information regarding the overall system and tend to determine anomalous logs from a local perspective, which is error-prone.
Besides, manually detecting anomalous log sequences is not efficient nor scalable because of the continuous increase in volume of logs generated by modern systems.

The use of keyword matching and regular expression helps to detect simple and well-known anomalous events.
Still, it is unable to identify a large portion of the anomalies,
as many of them are sequences of \textit{‘‘non-anomalous"} logs leading to an undesired outcome.
Automated log-based anomaly detection is an active research field.
Different approaches have emerged, driven by the need for robust and real-time solutions\cite{xu2009largescale,lou2010mining,lin2016log,du2017deeplog,zhang2019robust}.

Logs are semi-structured strings with a message field for free text. This format is convenient for human inspection,
but not for automated analysis. Existing log anomaly detection approaches enlighten the importance of 
structuring logs messages\cite{zhu2019tools}. 

3DS OUTSCALE\cite{outscale} is a French provider offering multi-sovereign Cloud services.
Logs are widely used within the company to monitor the different systems forming its Cloud computing platform.
Outside of monitoring, logs are also used to identify undesired outcomes such as virtual machine crashes or malicious security events.
In the case of this thesis, doing research on the detection of 
anomalies will improve operations and strengthen the high availability and security advocated by the company.

Within our Cloud computing platform, logging presents the following characteristics:
\begin{enumerate}
	\item Development teams use continuous integration\cite{fowler2006continuous} methods as a delivery model.
	It is a commonly used practice in software development and a great 	way to launch new features quickly. 
	But it also means that the code base and log statements evolve at a fast pace, which eventually induce instability within the log stream. \cite{kabinna2018examining}.
	 \item The spatial distance between log sources and the different storage systems is variable.
	 This configuration induces noise, as logs can arrive in mixed order or sometimes be duplicated.
\end{enumerate}

Regarding anomalies, some of them require a multi-source scope to be detected. For instance, certain patterns within storage logs are anomalous
only if certain actions are logged by network logs at the same time. 
This also highlights the importance of building robust, automated anomaly detection systems\cite{zhang2019robust}.

Our Ph.D. motivation is to bring forward MoniLog, an autonomous system for anomaly detection within logs, adapted to Cloud computing constraints
and large-scale online platforms in general. It is a robust and scalable way of parsing and analyzing logs, but also 
to classify the identified anomalies.

We based our design on a three-step approach presented in Section~\ref{sec:monilog}.  
Sections \ref{sec:anomaly_detection}, \ref{sec:logparsing}, and \ref{sec:anomaly_classification} are respectively 
dedicated to each step of our approach. Those sections present the state of the art and our planned contributions for our Ph.D.

Our work is highly motivated by the Cloud computing field limitations and needs, but those concerns exist in almost every large-scale, complex, online system.
Therefore, MoniLog does not aim to be limited to Cloud, but to be a robust solution for anomaly handling within constraint environments. Section \ref{sec:conclusion} 
sums up this article and expands the reflection on MoniLog applications.

\section{Monilog design}\label{sec:monilog}
\begin{figure}[h]
  \centering
        \includegraphics[width=\linewidth]{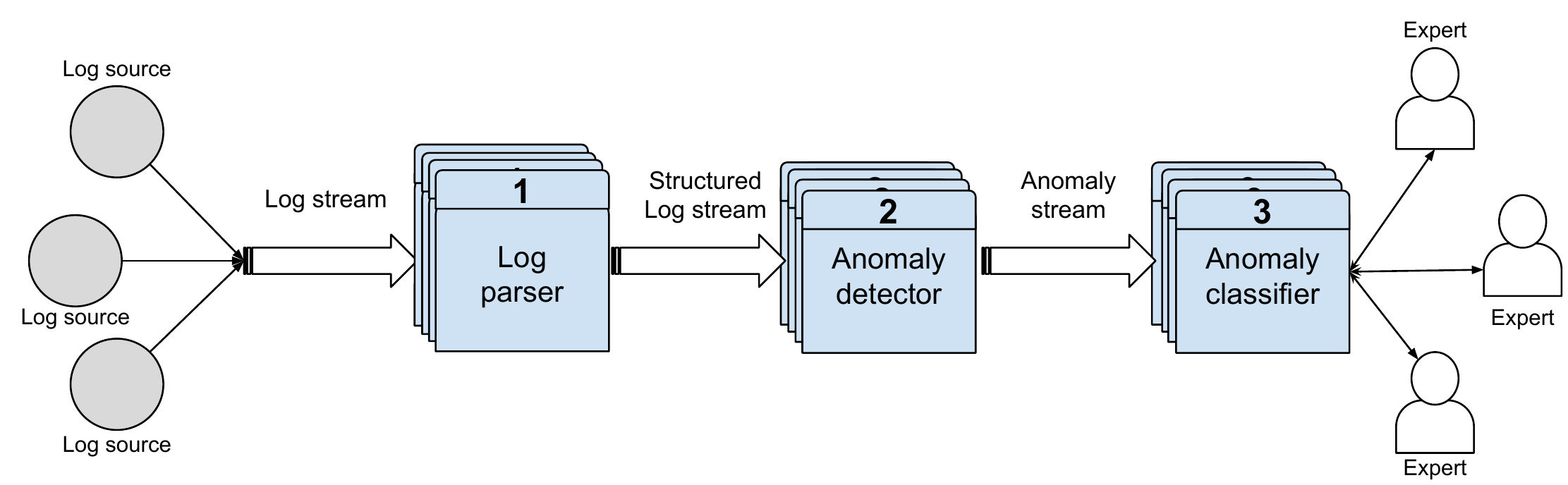}
\caption{MoniLog system design}
\label{fig:autonomous_design}
\end{figure} 

A large system is often composed of different softwares. Due to this amalgamated nature, it is linked to multiple log sources. 
At 3DS OUTSCALE, for instance, one system is connected to 24 different log sources and generates millions of log lines each second. 
To represent this, we modeled MoniLog input as a log stream fueled by various log sources. 

It is important for MoniLog components to be distributable in order to ensure scalability,
which is, for most of the online services companies, the core factor for growth.

The aimed output for MoniLog is a stream of classified anomalies with an assigned criticality.
To achieve this, we have designed a three-step system (Figure \ref{fig:autonomous_design}):
\begin{enumerate}
	\item A parsing component to extract logs' underlying pieces of information and obtain a structured log-stream. 
	\item A component to detect anomalies within a structured log stream. 
	This part generates anomaly reports, composed of all the logs linked to the identified anomalous sequence. 
	\item A classifier in charge of assigning anomalies a type and a level of criticality. 
	Anomalies types and criticality levels are defined by the monitoring teams. 
	This module is passively trained by observing the administrator's actions (Section \ref{sec:anomaly_classification}).
\end{enumerate}

\section{Log-based anomaly detection}\label{sec:anomaly_detection}

\begin{table}[!h] \renewcommand{\arraystretch}{1.3} \caption{Examples of log messages} \label{tab:logline_exemple}
\centering
\begin{tabular}{c|c}
\hline
\bfseries Id & \bfseries message\\ \hline\hline
\textit{$L_1$} & Sending 138 bytes src: 10.250.11.53 dest: /10.250.11.53\\
\textit{$L_2$} & Error while receiving data src: 10.250.11.53 dest: /10.250.11.53\\
\textit{$L_3$} & Sending 745675869 bytes src: 10.250.11.53 dest: /10.250.11.53\\
\textit{$L_4$} & Failed to verify data integrity src: 10.250.11.53 dest: /10.250.11.53\\
\hline
\end{tabular}
\end{table}

Programs are usually executed according to a fixed flow, and logs are produced according to those sequences.
Log-related anomalous events can be broadly divided into two categories:
\begin{enumerate}
	\item Sequential anomalies occur when a log sequence deviates from the normal flow (e.g., Table \ref{tab:logline_exemple}: \textit{$L_1$} $\rightarrow$ \textit{$L_4$} $\rightarrow$ \textit{$L_2$}).
	\item Quantitative anomalies are logs following the normal flow but with unusual values leading to an undesired outcome (e.g.,  \textit{$L_3$}).
\end{enumerate}

Existing solutions for anomaly detection within logs can be classified into two
categories: log message counter-based approaches (e.g., PCA\cite{xu2009largescale}, Invariant
Mining(IM)\cite{lou2010mining}, LogClustering\cite{lin2016log}) and deep learning-based approaches (e.g.,
 DeepLog\cite{du2017deeplog}, LogRobust\cite{zhang2019robust}, LogAnomaly\cite{meng2019loganomaly}).
 To compare them, we considered the following metrics:
 \begin{itemize}
 	\item \textit{Precision:} the percentage of log sequences correctly identified as anomalies within all the log sequences identified as anomalous by the model: $\textit{Precision}=\frac{TP}{TP + FP}$
	\item \textit{Recall:} the percentage of log sequences correctly identified as anomalies within all anomalous log sequences: $\textit{Recall}=\frac{TP}{TP + FN}$
	\item $\textit{F1-score}=\frac{2*\textit{Precision}*\textit{Recall}}{\textit{Precision} + \textit{Recall}}$
 \end{itemize}
 TP represents the number of abnormal log sequences that are correctly detected by the model, 
 FP the number of normal log sequences that are wrongly identified as anomalies by the model, and
 FN the number of abnormal log sequences that are not detected by the model.
 
The newest and more promising methods are based on deep-learning methods.
DeepLog, LogAnomaly, and LogRobust use Long-Short-Term Memory Neural Network (LSTM)\cite{schmidhuber1997long, huang2015bidirectional},
 a variant of the Recurrent Neural Network (RNN) model. 
LSTMs networks use a loop to forward the output of the last states to the current input.
Their structure is efficient to learn sequential patterns within the normal execution flow.

DeepLog uses a second LSTM to detect quantitative anomalies. It uses the knowledge of seen value to define if a new one is in the expected range.
LSTMs networks take a fixed-length vector as input. In DeepLog cases, this vector length represents the number of 
log patterns identified by log parsers on the training dataset. 
It works if one makes the closed-world assumption that log statements remain the same. 
It is not the case in our considered environments, where continuous integration may add, delete, or modify a log statement at any time. 

To deal with log template instability, LogRobust uses a log count vector to feed the neural network.
It is produced through a step called \textit{semantic vectorization}\cite{zhang2019robust},
in which semantic relationships between tokens are used to create fixed-length vectors.
This method is used to vectorize a new template without changing the vector length. 

To test their approach, LogRobust authors used different altered versions of an HDFS dataset.
 Each version contains a proportion from 0 to 20\% of unstable log events. 
Diverse types of events are crafted to represent real-world instabilities:
 \begin{itemize}
 	\item{Badly parsed logline, to simulate log parsing errors.}
	\item{Logline generated by twisting existing log statements to simulate a code modification.}
	\item{Duplicated or shuffled log to represent the noise and delay that can occur in a production environment.}
\end{itemize}

LogRobust is trained using a training set composed at 50\% by anomalous loglines. 
Creating a real-life dataset containing a lot of anomalies is complicated due to their rare nature, just as crafting or injecting anomalies into a training set is error-prone.
There is a risk that the structure learns how to recognize anomalous sequences instead of the normal flow, thus leading to degraded real-life performances of the system. 

LogAnomaly explores another option for facing log template instability. 
The authors intuition is that the majority of the new templates are just a minor variant of an existing one. 
To feed its LSTM structure, their system computes the similarity between a new template and the existing ones to find the best match. 

Regarding the presented state of the art and to design our detection component, we planned to conduct the following experiments:
\begin{itemize}
	\item To our knowledge, no deep comparative studies of LogAnomaly and LogRobust exist.
	We are interested in studying their precision if trained using an anomaly-free dataset.
	\item All the presented anomaly detection approaches use structured logs as input, and log parsing is not an error-free step (Section \ref{sec:logparsing}). 
	We want to evaluate the robustness of LSTM approaches regarding the potential errors due to the parsing step.
	\item LSTMs are good at learning sequences, but in a multi-source environment, execution flows from each source are mixed. 
	We want to compare LSTM with PCA, IM, and LogClustering approaches using a dataset extracted from such environment. 
\end{itemize}

\section{Log parsing}\label{sec:logparsing}

\begin{figure}[t]
  \centering
        \includegraphics[width=\linewidth]{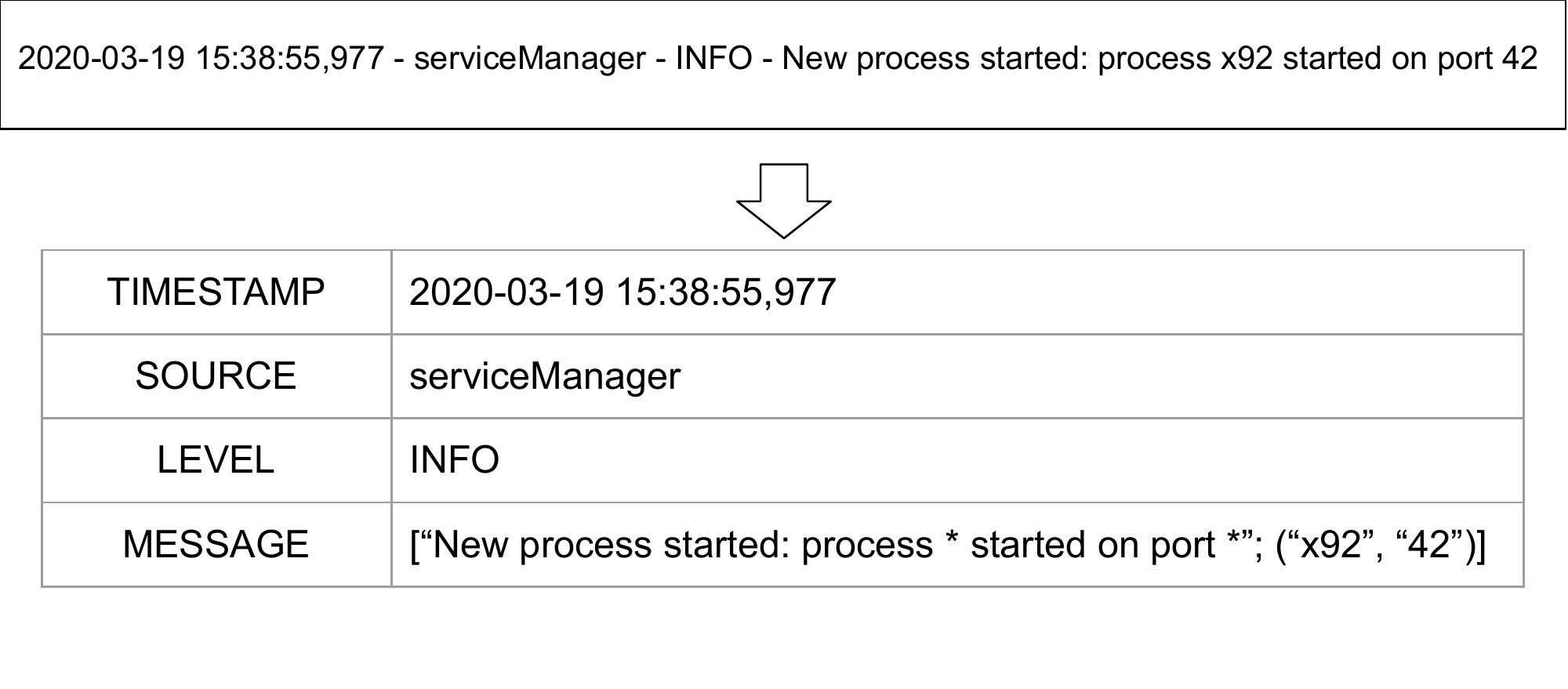}
\caption{Log parsing step}
\label{fig:log_parsing_exemple}
\end{figure}
A log can be divided into two parts:
\begin{itemize}
	\item A HEADER, composed of different fields such as timestamp, criticality level, source, etc.
	\item A MESSAGE, which is a text field without format constraint. Additional information can be added to this field by concatenating XML or JSON-formatted data with the free text.
\end{itemize}

The anomaly detection models previously presented used structured log data.
The HEADER fields are already structured according to a predefined format.
On the other hand, the MESSAGE field is composed of a static part (template) and of a variable part (variables).
The log parsing challenge lies within the discovery of those two parts (e.g., Figure \ref{fig:log_parsing_exemple}). 

Different algorithms exist in the literature, some of them process logs per batch (e.g., IPLoM\cite{ makanju2009clustering},SLCT\cite{vaarandi2003data}, LogCluster\cite{vaarandi2015logcluster}) 
and others process logs in a streaming fashion (e.g., Drain\cite{he2017drain}, Spell\cite{du2016spell}, Logan\cite{agrawal2019logan}, Logram\cite{dai2020logram}). 

On one hand, batch analysis has a major con regarding our context: Log statement instability made it impossible to collect a representative training set as it will never include yet non-existing log templates.
On the other hand, logs are produced streamingly and online parsing methods can discover new patterns on the job.
Moreover, online methods tend to have better results even on fixed datasets\cite{zhu2019tools}.
We would like our log parsing component to have some features:
\begin{enumerate}
	\item Being deployed without any human intervention 
	\item Processing logs in a streaming fashion
\end{enumerate} 
Regarding this, we therefore choose to concentrate our work on online methods.

To our knowledge, all the existing online log parsing algorithms use parameters, and most of them offer the possibility of pre-processing loglines.
 Fixing parameter values required a labeled dataset, which is costly and sometimes impossible.
 During the preprocessing step, algorithms use human crafted regular expressions to identify common variables such as URLs or IP addresses. 
 Preprocessing needs experts to define the regular expressions, which has a cost in time and can lead to mistakes impacting the parsing efficiency. 

According to recent studies\cite{zhu2019tools,zhang2019robust,du2017deeplog}, Drain is the most efficient existing parsing solution. 
We studied and experimented this approach and came up with two limitations to complete automation\cite{RNTI/papers/1002662}:
\begin{itemize}
  \item Drain's accuracy is influenced by preprocessing and can affect the accuracy if not well defined.
  \item Drain use two hyper-parameters and their values have a significant impact on precision. Therefore, Drain cannot be deployed in an unknown system with a high level of confidence.
\end{itemize}

By extending our study to other online log parsing solutions (Spell, Logram, Logan, SHISO\cite{mizutani2013incremental}, LenMa\cite{shima2016length}), 
we could like to present a benchmark of existing online log parsing approaches, focusing on their automation limits.

Regarding the distribution, Drain method, which show the best performances, is not distributable.
We plan to provide a distributed version of research tree-based log parsing method as we already have some encouraging results.

While working on logs coming from internal services, we noticed that almost 60\% of the tokens composing log messages are coming from JSON or XML-formatted data.
Adding such format at the end of a logline is a common practice within API-like services.
This indeed helps understand to understand the context of a log (e.g., "Send 42 bytes to 121.13.4.26 \{user\_id=125, service\_name=dart\_vader\}").
We therefore recommend a preliminary step to extract potential data coming from a structured format. This helps reduce the average length of log messages and can increase the discovery rate of
log parsing algorithms.

Current reference metrics to evaluate log parsing solutions are accuracy and time processing\cite{zhu2019tools, he2016evaluation}.
If we look at Table \ref{tab:logline_exemple}, log message \textit{$L_1$} \& \textit{$L_3$}  are considered correctly classified if they are identified as coming from the same log class.

The above metric is relevant when it comes to detect sequential anomalies as we are looking for uncommon log template sequences.
However, when it comes to quantitative anomalies, detection is only possible if the variable parts were correctly identified.
Therefore, we would like to propose a metric to evaluate wheter the static and variable parts of a log message are correctly identified.
A token is a sequence delimited by spaces inside a log message. 
In Table \ref{tab:logline_exemple}, log messages \textit{$L_1$} \& \textit{$L_2$} have respectively 7 \& 8 tokens.

\begin{equation}
\frac{1}{n}\sum_{i=1}^{n} \frac{1}{l_i}\sum_{j=1}^{l_i} \left\{ \,
\begin{IEEEeqnarraybox}[][c]{l?s}
\IEEEstrut
1 & if  $t_j = T_j$ \\ 
0 & if  $t_j \neq T_j$ 
\IEEEstrut
\end{IEEEeqnarraybox}
\right.
\label{eq:parsing_ganular}
\end{equation}

Considering a pool of n parsed loglines, $l_i$, represents the number of tokens within logline i, $t_j$ the value of the j-th token (static or variable), and $T_j$ the expected value of the j-th token.

Evaluating existing log parsers with this metric will give us a better comprehension of their capacity to extract variables from log messages and their relevance for detecting quantitative anomalies.

Most of the existing metrics to evaluate log parsing methods performances are supervised.
Within their papers presenting Logan, the authors introduce an unsupervised metric\cite{agrawal2019logan}.
We plan to extend that study to the pertinence of other unsupervised metrics\cite{palacio2019evaluation}. 

Unsupervised metrics opens promising perspectives for auto-parametrizing log parser. 
We can imagine a component deployed according to the following flow.
First, it acquires a fixed quantity of loglines within its environment.
Then it calibrates the value of its parameters by estimating its performance using an unsupervised metric. 
Once it detects the supposed optimal values, it starts parsing logs. 

\section{Classify anomalies}\label{sec:anomaly_classification}

\begin{figure}[t]
  \centering
        \includegraphics[width=0.9\linewidth]{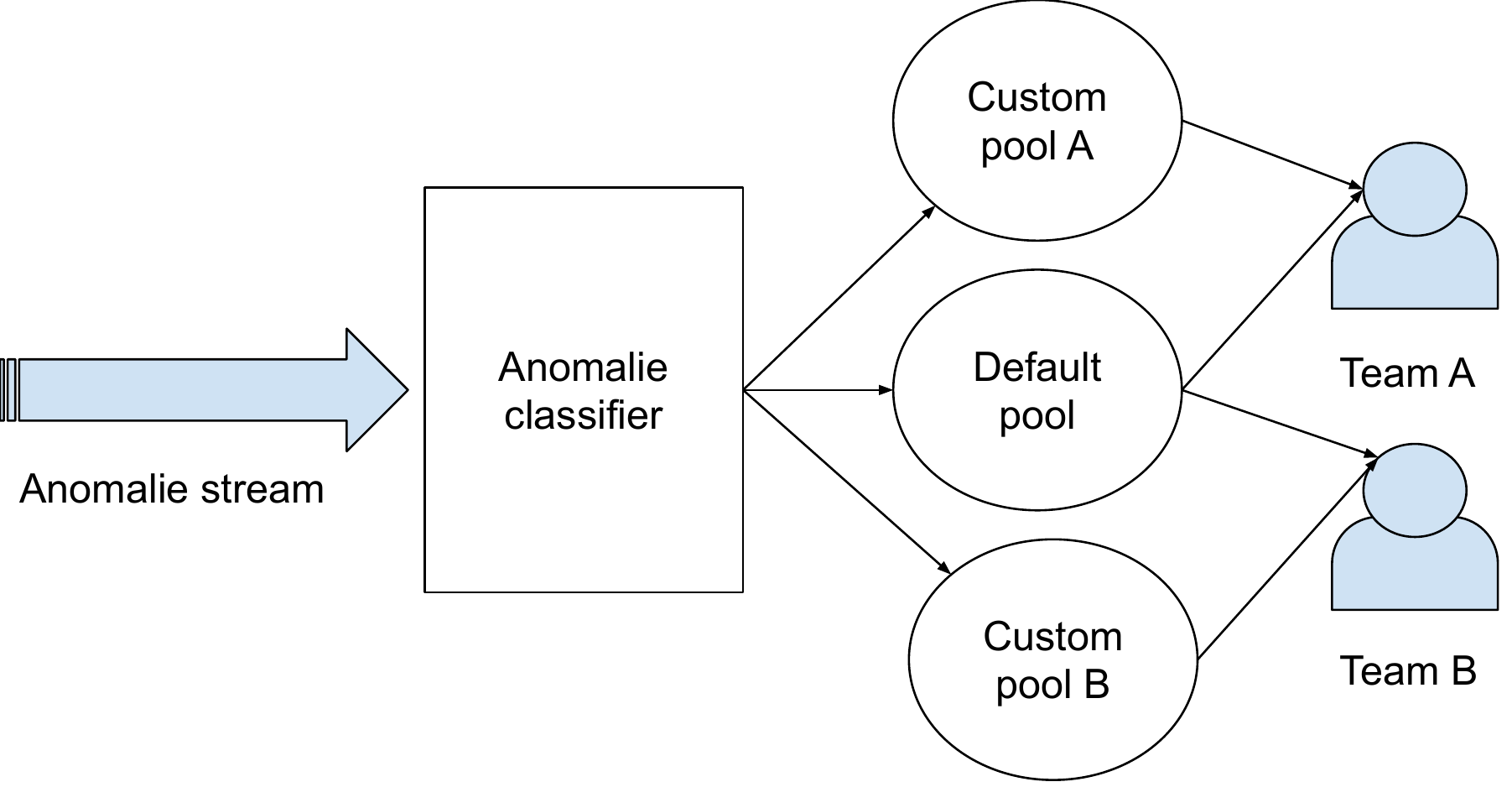}
\caption{MoniLog system design}
\label{fig:anomaly_classifier}
\end{figure} 

Anomalies can be different types of events, from breakdowns to performance issues, including security attacks.
Due to their different natures, anomalies are generally not handled by the same team and with the same priority.
A common practice to prioritize the tasks is to assign anomalies a level of criticality such as \textit{“Low"},  \textit{“moderate"} or \textit{“high"}. 

To our knowledge, few studies exist for automated anomaly classification within logs.
Meng \& al. propose LogClass, trained a classifier over log anomalies\cite{meng2018device}.

Everyone can define a set of log class and a criticality scale appropriate for their domain.
This subjective nature motivates us to present a customizable and intelligent module
for anomaly classification (Figure \ref{fig:anomaly_classifier}).

We would like to provide an easy way for monitoring teams to assign the resolution tasks.
We plan our component to work using a pool system.
Initially, there is just one default pool, but additional pools can be created or deleted by administrators. 

Each time an alert is moved from a pool to another, it is used as an assessment signal to enrich the algorithm's ability to classify further anomalies within a specific pool.
In the same way, every time the level of criticality is manually modified, it is used to improve further anomaly evaluation.

We believe that this flexible design will allow maintainers to reproduce their anomaly handling process. 
This is also a convenient way to provide feedback to the classifier without any extra human effort as it is passively done by the user experience.
\section{Conclusion}\label{sec:conclusion}
In this article, we presented MoniLog, an automated approach for log-based anomaly detection within Cloud computing infrastructures.
We based our planned system design on three distributable and therefore easy-scalable components.
The first one parses the log message to extract relevant pieces of information.
The produced structured stream is analyzed in order to identify anomalous sequences.
Once detected, anomalous events are assigned a criticality level and a class.

MoniLog was designed to be robust to log statement evolutions.
It also allows real-time scalable anomaly detection within complex logging environments.
Even if it were conceived for Cloud computing, we believe the approach to be relevant for any large-scale online services.

I would like to thank Professor Raja Chiky for her guidance, encouragement, and valuable advice.
A special mention to Mar Callau-Zori for her time and precious feedbacks.
My thanks also go to 3DS OUTSCALE and the ANRT, which have funded my thesis and gave me access to real-life data, computing resources, as well as the opportunity to exchange with field experts daily. 
\balance

\bibliographystyle{IEEEtran}
\bibliography{biblio}

\end{document}